# ACQUISITION ACCURACY EVALUATION IN VISUAL INSPECTION SYSTEMS - A PRACTICAL APPROACH

Radu Arsinte
Software ITC - Hardware Department
Str.Gh.Bilascu 109 , 3400 Cluj-Napoca - Romania
tel. +40-64-197681,197682 ext.28 - fax.+40-64-191236,196787
Costin Miron
Technical University Cluj-Napoca - Faculty of Electronics and Telecommunications
Str.Gh.Baritiu 26-28 ,3400 Cluj-Napoca,Romania

*Abstract* -This paper draws a proposal of a set of parameters and methods for accuracy evaluation of visual inspection systems. The case of a monochrome board is treated ,but practically all conclusions and methods may be extended for colour acquisition. Basically, the proposed parameters are grouped in five sets as follows:Internal noise;Video ADC cuantisation parameters;Analogue processing section parameters;Dominant frequencies;Synchronisation (lock-in) accuracy. On basis of this set of parameters was developed a software environment , in conjunction with a test signal generator that allows the "test" images. The paper also presents conclusions of evaluation for two types of video acquisition boards.

## I. INTRODUCTION

Processing algorithms, as used in current visual inspection applications are sometimes very sensitive to a change of concrete conditions: illumination, resolution of input image , noise level of the image. Many of those parameters are dependent on the image acquisition hardware .Classical parameters (spatial resolution, acquisition clock) does not offer enough information about metrologic accuracy of measurement, which can affect processing algorithm's stability. Despite this fact, few papers are treating those concrete problems , and no one has proposed a set of parameters and methods for the right evaluation.

The simplest way to implement a test system is to use a generator of image-like waveforms (video generator) , in conjunction with the image acquisition system being tested.

## II. DEFINITIONS OF PARAMETERS

**1.Internal Noise -**Noise is a general propriety of electronic systems. The mathematical theory is generally well known [1], but identifying and measuring it is more difficult .We propose in this part a couple of parameters , enough for comparison purposes in most cases .Measurement can be done in a simple system , applying from the video generator a specific waveform (Waveform1-fig.1) or a "uniform grey" image obtained by TV camera's lens obturation .Combining the two methods the error contribution of TV camera may be deduced.

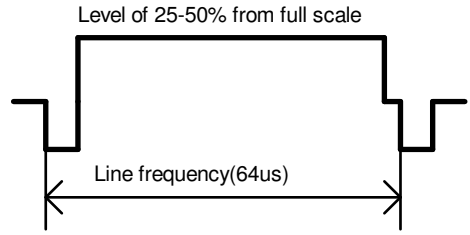

Fig.1.Waveform used for the noise performance evaluation

The proposed parameters are:
a).**Absolute mean value** of noise defined as follows ([ 2 ] [ 4 ]):

$$Na = \frac{\sum_{i=0}^{N-1}\sum_{j=0}^{N-1}(|e(i,j)-M|)}{NxN} \qquad (1)$$

where :
M - mean value of the "dark" image computed with the following formula:

$$M = \frac{\sum_{i=0}^{N-1}\sum_{j=0}^{N-1}(e(i,j))}{NxN} \qquad (2)$$

e(i, j) -grey value of pixel located in co-ordinates (i,j)
N - spatial resolution of the acquisition section (for simplicity reasons ,resolutions on x and y-axis are considered the same)
b).**Maximum value** of noise :

$$N\max = \max_{i,j}^{N-1}(|e(i,j)-M|) \qquad (3)$$

c).An other useful parameter is the average **mean-square value** of noise defined as[1]:

$$N_{ms}^2 = \frac{\sum_{i=0}^{N}\sum_{j=0}^{N}N(e(i,j)-M)^2}{NxN} \qquad (4)$$

d).Experimentally, the average mean square error is estimated by the **average sample mean-square** error in the given image:

$$N_{ms}^2 \cong \frac{\sum_{i=0}^{N}\sum_{j=0}^{N}(e(i,j)-M)^2}{NxN} \qquad (5)$$

An other method of evaluation is based on histogram , measuring the surface of the peak around the medium level of grey.

**2.A/D converter cuantisation parameters** -A frame-grabber generally uses a flash ADC , with a sampling frequency exceeding 10-15 Msps. Although a large offer of such high performance converters exists, many producers don't offer any guarantees of monotonicity , or missing codes. Evaluation of ADC used in inspection system, even not complete [ 3 ] is important .

Evaluation can be made in the same manner as for the noise parameters. In this case we can apply Waveform 2 (fig.2).

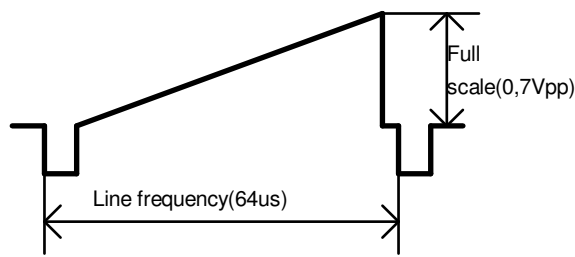

Fig.2.Waveform used for ADC parameters evaluation

The file resulted from acquisition, after applying this waveform can be further processed, building the grey level histogram. From the histogram missing codes can be easy revealed, corresponding position from histogram being zero or significantly reduced compared to the others' positions. This test can also reveal the real resolution of the ADC used.

ADC parameters can be also evaluated in a simple manner, processing a "real image" file. In this case the choosen image must contain as much is possible grey levels or several files can be used for the right evaluation and error minimisation.

**3. Analogue processing section parameters -** The evaluation of parameters involved in analogue processing section is extremely difficult. The reason is the fact that this section has performances adequate of video signal processing (large bandwidth, short rise and fall time).

*a).Black level restoration stability*- in most frame -grabbers black level (reference level) of the input image is restored using a clamp circuit, mainly composed from an analogue switch and a storage capacitor. Usually, the black level is restored at the beginning of every information line, by opening the switch. Due of the transient processes this level is different from line to line, this difference being not greater than 1-2 LSB. Evaluation of this parameter is done in the same configuration used for noise measurement (fig.1).

*b).Black level decay*-The charge of the storage capacitor decreases in time, causing a so call "level decay". Usually this loss of voltage, for an information line, is less than 1LSB, but in certain cheap boards may be 4-5 times greater. A combined evaluation of these two aspects can be performed for a "uniform grey" image (as discussed at pct.1).Practically the image is divided in equal squares (elementary surfaces-8x8 pixels or 16x16 pixels) for each of them calculating a mean value as follows:

$$M(m,n) = \frac{\sum_{i=0}^{15}\sum_{j=0}^{15} e(m*16+i, n*16+j)}{256} \quad (6)$$

Obs. A 16x16 pixel of elementary surface is assumed

For a high performance board ,differences between the left-up corner and right-down corner of the matrix must be not greater than 2LSB's.

*c).Rise and fall times-*this test can offer a measure of bandwidth of input amplifier applying a waveform as described in fig.3.Practically the lenght of every transition (exprimated in pixels) is measured.

**4. Dominant frequencies-**In every image acquisition system, due to the particular configuration, components are influencing each other. In general manner, this influence is included in "noise parameters" as discussed in cap.1. This paragraph is focused on those "noise" components that have a regular structure: harmonic frequencies , radiated perturbations.

In a particular system , a PC for example, perturbations are caused by the power supply and clock oscillators of different boards from the  system.

Evaluation of those perturbations, here called **"dominant frequencies"** is obtained by FFT analysis ([3],[4],[5])of every line or column. The processed file is the same as used in #1 (noise measurement).

Considering Fk(i) the FFT power coefficients for the k information line ( i varying from 0 to N/2) the global coefficient Fg(i) is calculated as following [ 3 ]:

$$Fg(i) = \sum_{k=0}^{N-1} Fk(i) \qquad (7)$$

From the set of Fg(i) values we can estimate if one frequency is dominant. Criteria for this evaluation can be a threshold level estimated from the media of Fg(i) elements:

$$Mf = \frac{\sum_{i=0}^{N-1} Fg(i)}{N} \qquad (8)$$

**5. Synchronisation accuracy-** Most frame grabber offers a so call "genlocking" facility, adjusting the internal clock (acquisition clock) using for reference the sync impulses of the video source. Usually this feature is provided by an PLL circuit. In this case a "jitters" between the video source clock and the frame grabber clock causes a "tooting" of vertical edges .This fact is less important in multimedia applications , visualisation applications  and far more important in image analysis, in particular in industrial visual sensors. Measurement of "jitters" is done usually using a scope or logic analyser, and value's offered(not often) by the manufacturers are 0.2-1 pixel. We describe a statistical method of evaluation using only the global system and statistical processing of measurements. Evaluations of the so call "Syncronisation accuracy" is done using the system described in fig.1. and a special waveform (Waveform 3-fig.4.).

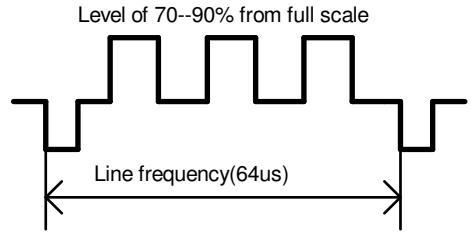

Fig.3.Waveform used for Synchronisation accuracy test

We call the coordinates of the line image memory corresponding to the fall and rise fronts **transition points**. In the ideal case, transition points for every line of information have the same value. Assuming that the Q transition points for the k line of information are:

$$m_1(k), m_2(k), ..., m_Q(k) \tag{9}$$

we consider as a measure of synchronisation accuracy the following formula:

$$Sy = \frac{1}{Q}\sum_{q=1}^{Q} \frac{\sum_{k=0}^{N-1}|m_q(k) - M_q|}{N} \tag{10}$$

Where Mq is the mean value of the q transition point computed as follows:

$$M_q = \frac{\sum_{k=0}^{N-1} m_q(k)}{N}; q = 1, 2, ..., Q \tag{11}$$

### III. IMPLEMENTATION AND EXPERIMENTAL RESULTS

All this set of methods was implemented in a software environment called **FGEVAL.** The main screen of the Windows version is presented in fig.4. As a video waveform generator can be used a VGA controller, reprogramming the internal registers to match the TV standard. The evaluation was made for several types of commercial boards and for one type of video acquisition board designed by the author. Performance evaluation obtained match in all cases subjective parameters (image "look") and in most cases offers a supplement of information. Parameters measured for two types of board are presented in table 1.

**Table 1.**

| Board | Noise performance | Black level stability | Dominant freq. | ADC parameters | Sync. parameters |
|---|---|---|---|---|---|
| No #1 | abs.< 1LSB RMS<1LSB | Variation <0.07%` | fs/8-75.5% fs/4-80% fs/2-100% 3fs/4-80% | No missing codes. | 0.72points/transition |
| No #2 | abs.<1.5LSB RMS <1.8 LSB | Variation <2% | - | No missing codes | 1.28 points /transition |

Note: fs-is the sample frequency specific for each type of board

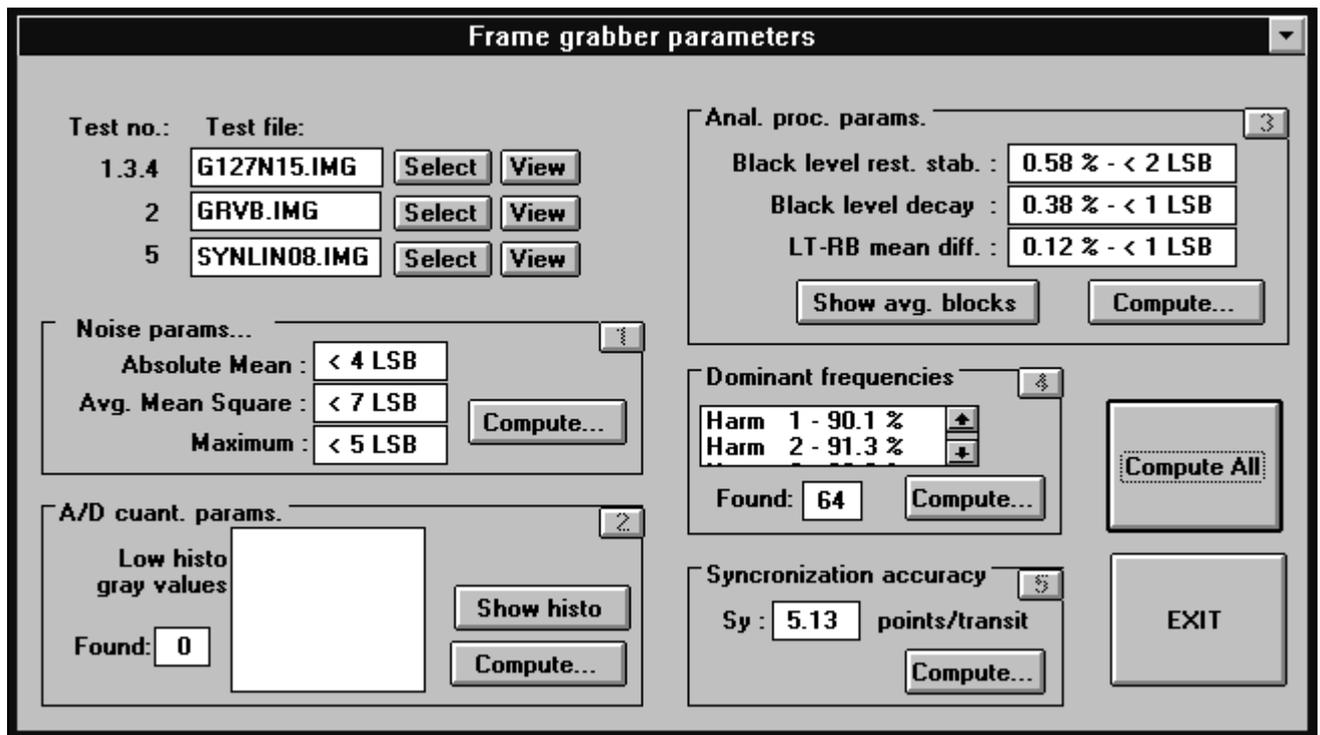

Fig.4.Main menu of the FGEVAL environement

## IV. CONCLUSION

Although frame-grabbers are common components of modern technology, testing the performances for a right evaluation, is difficult and rarely treated. A simple system, as described in this paper (an enhanced version of [6]), combined with methods and software processing tools can be a real help for current purposes.